\documentclass[preprint,12pt]{elsarticle}

\usepackage{amssymb}
\usepackage{multirow}
\usepackage{hyperref}
\usepackage{lineno}
\usepackage{float} 
\usepackage{xcolor}
\linenumbers

\begin{document}
\nolinenumbers
\begin{frontmatter}



\title{IndoHerb: Indonesia Medicinal Plants Recognition using Transfer Learning and Deep Learning}


\author[inst1]{Muhammad Salman Ikrar Musyaffa'}

\affiliation[inst1]{organization={Informatics Engineering, Faculty of Computer Science, Brawijaya University, Malang, 65145, East Java, Indonesia}
}

\affiliation[inst2]{organization={Depatement of Biology, Faculty of Mathematocs and Natural Science, Brawijaya University, Malang, 65145, East Java, Indonesia}}

\author[inst1]{Novanto Yudistira}
\author[inst1]{Muhammad Arif Rahman}
\author[inst2]{Jati Batoro}


\begin{abstract}

The rich diversity of herbal plants in Indonesia holds immense potential as alternative resources for traditional healing and ethnobotanical practices. However, the dwindling recognition of herbal plants due to modernization poses a significant challenge in preserving this valuable heritage. The accurate identification of these plants is crucial for the continuity of traditional practices and the utilization of their nutritional benefits. Nevertheless, the manual identification of herbal plants remains a time-consuming task, demanding expert knowledge and meticulous examination of plant characteristics. In response, the application of computer vision emerges as a promising solution to facilitate the efficient identification of herbal plants. This research addresses the task of classifying Indonesian herbal plants through the implementation of transfer learning of Convolutional Neural Networks (CNN). To support our study, we curated an extensive dataset of herbal plant images from Indonesia with careful manual selection. Subsequently, we conducted rigorous data preprocessing, and classification utilizing transfer learning methodologies with five distinct models: ResNet, DenseNet, VGG, ConvNeXt, and Swin Transformer. Our comprehensive analysis revealed that ConvNeXt achieved the highest accuracy, standing at an impressive 92.5\%. Additionally, we conducted testing using a scratch model, resulting in an accuracy of 53.9\%. The experimental setup featured essential hyperparameters, including the ExponentialLR scheduler with a gamma value of 0.9, a learning rate of 0.001, the Cross-Entropy Loss function, the Adam optimizer, and a training epoch count of 50. This study's outcomes offer valuable insights and practical implications for the automated identification of Indonesian medicinal plants, contributing not only to the preservation of ethnobotanical knowledge but also to the enhancement of agricultural practices through the cultivation of these valuable resources. The Indonesia Medicinal Plant Dataset utilized in this research is openly accessible at the following link: \href{https://github.com/Salmanim20/indo\_medicinal\_plant}{https://github.com/Salmanim20/indo\_medicinal\_plant}.
\end{abstract}


\end{frontmatter}


\section{Introduction}
\label{sec:sample1}

Indonesia, with its tropical climate and abundant biodiversity, is home to a vast array of plant species, among which herbal plants hold a prominent position. This archipelagic nation boasts approximately 30,000 distinct plant species, with around 9,600 classified as herbal plants. These herbal plants have been used for generations as a means of traditional healing. However, with ongoing societal and technological advancements, the use of herbal remedies has gradually declined, as modern pharmaceuticals increasingly supplant traditional practices. This transition, while facilitating access to advanced medical care, threatens to overshadow the long-established efficacy of herbal treatments. Therefore, it is crucial to explore innovative approaches that revive the recognition and use of herbal plants, especially in regions where access to modern medicine is limited by economic or geographical constraints.

The significance of herbal plants extends beyond their therapeutic potential. In an era marked by a heightened awareness of health and a growing demand for natural, chemical-free food products, herbal plants have found renewed importance. They play a pivotal role in the production of organic, health-conscious foods and are central to a natural, holistic lifestyle. This resurgence of interest in herbal plants has led to a surge in organic cultivation practices, aligning with the broader trend of environmental sustainability \cite{Yulianto2017}.

Accurate identification of herbal plants is essential, but it presents a formidable challenge, often demanding an exhaustive understanding of plant phenotypes and intricate botanical knowledge. Plant identification encompasses a multitude of criteria, including color, flower morphology, leaf structure, texture, and overall plant architecture. The vast number of plant species, coupled with morphological similarities among closely related species, compounds this challenge \cite{Zin2020}. The advent of computer vision technology in recent years offers a promising solution to expedite and enhance the accuracy of plant identification. Utilizing integrated cameras and machine learning algorithms, individuals can access plant information swiftly and accurately. Consequently, computer vision has become a focal point of research in the domain of herbal plant identification, leveraging various plant features such as leaves, roots, and fruits \cite{Quoc2021}.

Previous research by Quoc and Hoang \cite{Quoc2021} explored herbal plant identification using the Scale-Invariant Feature Transform (SIFT) \cite{Lindeberg2012} and Speeded Up Robust Features (SURF) \cite{Bay2006} algorithms. The study employed two resolution versions, 256x256 and 512x512, yielding accuracy rates ranging from 21\% to 37.4\%. Additionally, Liantoni \cite{Liantoni2015} investigated classification methods, including Naive Bayes \cite{Webb2017} and K-Nearest Neighbor \cite{Seidl2009}, achieving accuracy rates of 70.83\% to 75\%. While these results provide valuable insights, they underscore the need for further improvement in classification accuracy.

This research aims to address the accuracy limitations observed in prior studies by adopting a transfer learning approach with Convolutional Neural Networks (CNNs) \cite{DBLP:journals/corr/OSheaN15}. Transfer learning offers a compelling advantage by combining feature extraction and classification algorithms, optimizing learning efficiency. Its proficiency in object classification within images makes it a promising candidate for enhancing prediction accuracy. Additionally, this study seeks to augment the research landscape by assembling a dedicated dataset of Indonesian herbal plants, meticulously collected through Google Images searches. {\color{black}However, it is important to acknowledge the limitations of this research. These limitations include the utilization of only five pre-trained models: ResNet, DenseNet, VGG11, ConvNeXt, and Swin Transformer. Additionally, the study focused exclusively on 100 classes, selected based on information obtained from an online source. Furthermore, the Dynamic Learning Rate scheduler used was restricted to the ExponentialLR scheduler.}

In conclusion, our contribution can be summarized into 3 folds:
\begin{enumerate}
  \item The adoption of the transfer learning approach involving CNNs to herbal plant identification.
  \item The creation of a dedicated dataset of Indonesian herbal plants, which is a valuable resource for future studies in the field of herbal plant identification.
  \item Advancement of herbal plant identification through the application of cutting-edge technology, fostering a renewed appreciation for the invaluable wealth of botanical resources in Indonesia and beyond.
\end{enumerate}

\section{Related Works}
Prior research in the field of medicinal plant classification has laid a significant foundation for our current study. Notable contributions include the work of Quoc \& Hoang \cite{Quoc2021}, who conducted extensive research on Vietnam Medicinal Plants. Their study employed the Scale-Invariant Feature Transform (SIFT) \cite{Lindeberg2012} and Speeded Up Robust Features (SURF) \cite{Bay2006} algorithms and featured a dataset comprising 20,000 herbal plant images. The research meticulously explored two resolution versions, 256x256 and 512x512, with varying accuracy rates. Specifically, in the 256x256 version, the SURF method achieved an accuracy of 21\%, while the SIFT method reached 28\%. In the higher-resolution 512x512 version, the SURF method improved to an accuracy of 34.7\%, and the SIFT method achieved 37.4\%.

Liantoni \cite{Liantoni2015} delved into herbal leaf classification, utilizing the Naive Bayes \cite{Webb2017} and K-Nearest Neighbor \cite{Seidl2009} methods. Their dataset comprised 120 images, with 96 images allocated for training and 24 for testing. The study revealed the superiority of the Naive Bayes method, achieving an accuracy rate of 75\% compared to the K-Nearest Neighbor method's 70.83\%. Both methods underwent rigorous testing, spanning 100 epochs.

Naeem et al. \cite{Naeem2021} contributed to the field with their focus on the classification of medicinal plant leaves. Their dataset included 6,000 leaf images distributed across six classes, with each class containing 1,000 images. The study employed an array of classification algorithms, such as Multi-Layer Perceptron (MLP) \cite{Almeida1997}, LogitBoost (LB) \cite{Friedman2000}, Bagging (B) \cite{breiman96}, Random Forest (RF) \cite{ho1995random}, and Simple Logistic (SL) \cite{Peng2002}. Two image sizes, 220x220 and 280x280, were considered, resulting in accuracy rates ranging from 92.56\% to 99.01\%. Notably, the Multi-Layer Perceptron method outperformed other methods, asserting its effectiveness in medicinal plant leaf classification.

Drawing inspiration from these notable research contributions, our current study aims to extend the existing body of knowledge by introducing a new dataset of Indonesian medicinal plants. This dataset, curated independently using the Google Images search engine, forms the basis for our exploration of transfer learning methods from convolutional neural networks to enhance classification accuracy. The challenges and strategies encountered during dataset collection and preparation will inform the robustness of our forthcoming research.

In summary, the related works discussed here provide valuable insights and benchmarks in medicinal plant classification, setting the stage for our contributions to this evolving field through the creation of a new dataset and the application of advanced machine learning techniques.

\section{IndoHerb Dataset}
\label{sec:sample2}

Based on the previous research, the aim of this study is to create a new dataset of medicinal plants from Indonesia, collected independently through the Google Images search engine called IndoHerb. {\color{black} To ensure that the plant images correspond to the classification, we cross-referenced the medicinal plants with the list provided by the Medicinal Plant Maintenance Installation at the Baturaja Health Research and Development Center, Indonesian Ministry of Health \cite{APAWeb}. Subsequently, the images were systematically collected from the web. Finally, the dataset underwent validation and meticulous selection by a professor of biology—an expert in Plant Taxonomy and Ethnobotany—to authenticate the collected images of Indonesian medicinal plants.} The datasets will be tested using the transfer learning method from the convolutional neural network.

During the dataset collection process, the initial step involved determining which classes of medicinal plants from Indonesia would be utilized for research. To identify these classes, searches were conducted on various websites that provided the names of medicinal plant classes from Indonesia, resulting in the collection of 100 classes for research. Following this, an image search was conducted for each class of medicinal plants using the Google Images search engine. 

\begin{figure}[H]
    \centering
    \includegraphics[width=13cm]{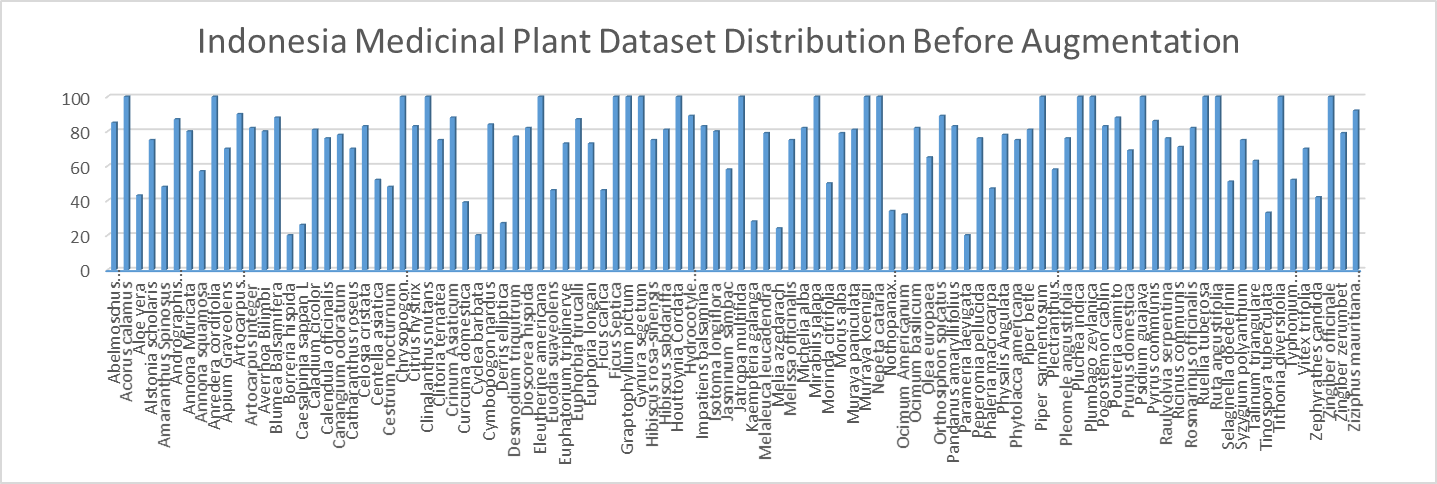}
    \caption{Graph of Indonesia Medicinal Plant Dataset Distribution Before Augmentation}
    \label{distribute}
\end{figure}

{\color{black}From the search results, images that met the criteria for each class were manually selected from the top search results. This manual selection was necessary because, at times, there were discrepancies, such as instances where identical images appeared from different sources. While this issue was minimized in the study, inadvertent selections of identical images were left unchanged. Additionally, some images were excluded due to quality issues, such as scribbles or watermarks. Another challenge involved less-known or rarely found classes, resulting in a limited number of search results (e.g., approximately 30 images meeting the criteria). To address this, an image augmentation process was implemented on the obtained images, including horizontal flips, vertical flips, and rotations. Ultimately, all images were resized to 128x128. The final dataset of medicinal plants from Indonesia comprises 10,000 images across 100 classes, with each class containing 100 images. Figure \ref{distribute} illustrates the dataset distribution before the augmentation process.}

Based on Figure \ref{distribute}, it can be seen that there are various numbers of images collected from each class before the augmentation process is carried out. This is due to the limited number of images that can be selected for research due to several problem factors as previously described. The class used in the dataset are Abelmoschus esculentus, Acorus calamus, Aloe vera, Alstonia scholaris, Amaranthus Spinosus, Andrographis paniculata, Annona Muricata, Annona squamosa, Anredera cordifolia, Apium Graveolens, Artocarpus heterophyllus, Artocarpus integer, Averrhoa Bilimbi, Blumea Balsamifera, Borreria hispida, Caesalpinia sappan L, Caladium cicolor, Calendula officinalis, Canangium odoratum, Catharanthus roseus, Celosia cristata, Centella asiatica, Cestrum nocturnum, Chrysopogon Zizanioides, Citrus hystrix, Clinalanthus nutans, Clitoria ternatea, Crinum Asiaticum, Curcuma domestica, Cyclea barbata, Cymbopogon nardus, Derris elliptica, Desmodium triquitrum, Dioscorea hispida, Eleutherine americana, Euodia suaveolens, Euphatorium triplinerve, Euphorbia tirucalli, Euphoria longan, Ficus carica, Ficus Septica, Graptophyllum pictum, Gynura segetum, Hibiscus rosa-sinensis, Hibiscus sabdariffa, Houttoynia Cordata, Hydrocotyle Sibthorpioides, Impatiens balsamina, Isotoma longiflora, Jasminum sambac, Jatropa multifida, Kaempferia galanga, Melaleuca leucadendra, Melia azedarach, Melissa officinalis, Michelia alba, Mirabilis jalapa, Morinda citrifolia, Morus alba, Muraya paniculata, Murraya koenigii, Nepeta cataria, Nothopanax scutellarium, Ocimum Americanum, Ocimum, basilicum, Olea europaea, Orthosiphon spicatus, Pandanus amaryllifolius, Parameria laevigata, Peperomia pellucida, Phaleria macrocarpa, Physalis Angulata, Phytolacca americana, Piper betle, Piper sarmentosum, Plectranthus Scutellarioides, Pleomele angustifolia, Pluchea indica, Plumbago zeylanica, Pogostemon cablin, Pouteria caimito, Prunus domestica, Psidium guajava, Pyrrus communis, Raulvolvia serpentina, Ricinus communis, Rosmarinus officinalis, Ruellia tuberosa, Ruta angustifolia, Selaginella doederlinii, Syzygium polyanthum, Talinum triangulare, Tinospora tuberculata, Tithonia diversifolia, Typhonium flagelliforme, Vitex trifolia, Zephyrathes candida, Zingiber officinale, Zingiber zerumbet, and Ziziphus Mauritiana Lam.

\begin{figure}[H]
    \centering
    \includegraphics[width=13cm]{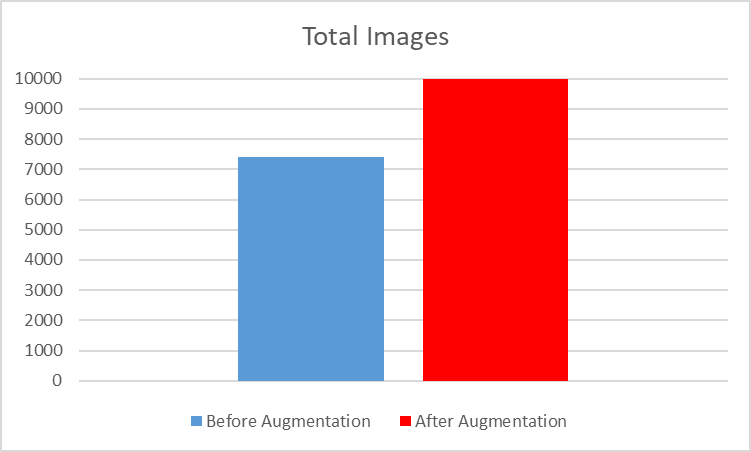}
    \caption{\color{black}{Total Images Before and After Augmentation}}
    \label{total_images}
\end{figure}

{\color{black}
Based on Figure \ref{total_images}, it can be seen that the number of Indonesian Medicinal Plant datasets that have been collected is 7,391 images. For this reason, an augmentation process is carried out for each class that do not meet the target number of 100 so that each class has the same number. In the end, we get 10,000 images in total where the 2,609 images are the result of the augmentation process in the form of horizontal flip, vertical flip, and rotation.
}
\section{Methodology}
\subsection{Gathering Data}
\begin{figure}[H]
    \centering
    \includegraphics[width=1.5cm]{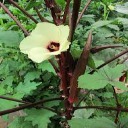}
    \includegraphics[width=1.5cm]{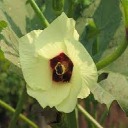}
    \includegraphics[width=1.5cm]{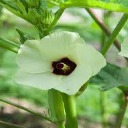}
    \includegraphics[width=1.5cm]{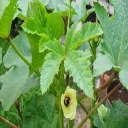}   
    \includegraphics[width=1.5cm]{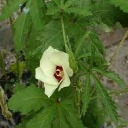} 
    \\Abelmoschus Esculentus\\
    \includegraphics[width=1.5cm]{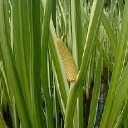}
    \includegraphics[width=1.5cm]{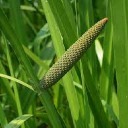}
    \includegraphics[width=1.5cm]{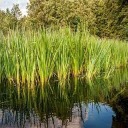}
    \includegraphics[width=1.5cm]{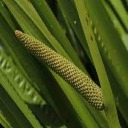}   
    \includegraphics[width=1.5cm]{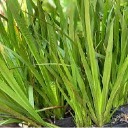}
    \\Acorus Calamus\\
    \includegraphics[width=1.5cm]{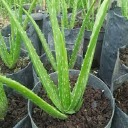}
    \includegraphics[width=1.5cm]{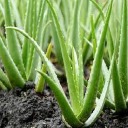}
    \includegraphics[width=1.5cm]{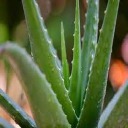}
    \includegraphics[width=1.5cm]{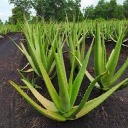}   
    \includegraphics[width=1.5cm]{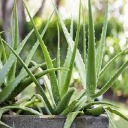}
    \\Aloe Vera\\
    \includegraphics[width=1.5cm]{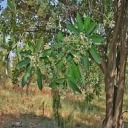}
    \includegraphics[width=1.5cm]{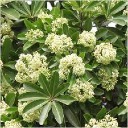}
    \includegraphics[width=1.5cm]{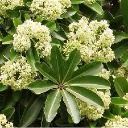}
    \includegraphics[width=1.5cm]{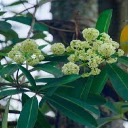}   
    \includegraphics[width=1.5cm]{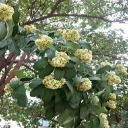}
    \\Alstonia Scholaris\\
    \includegraphics[width=1.5cm]{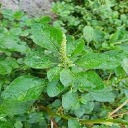}
    \includegraphics[width=1.5cm]{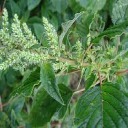}
    \includegraphics[width=1.5cm]{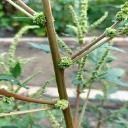}
    \includegraphics[width=1.5cm]{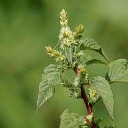}   
    \includegraphics[width=1.5cm]{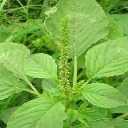}
    \\Amaranthus Spinosus\\    
    \caption{Example Images From Five Selected Classes of Indonesia Medicinal Plant Dataset}
    \label{example_data}
\end{figure}

The collected data consists of images of herbal plants sourced from two main datasets: the Vietnam Medicinal Plant public dataset and the Indonesia Medicinal Plant dataset. The Vietnam dataset comprises a total of 20,000 images categorized into 200 classes. On the other hand, the Indonesia Medicinal Plant dataset was obtained independently through Google Images search engine and includes 10,000 images from 100 classes. Example images from the Indonesia Medicinal Plant Dataset are illustrated in Figures \ref{example_data}.

Based on Figure \ref{example_data}, it showcases examples of 5 classes from the Indonesian herbal plant dataset, each consisting of 5 image examples. These images were collected independently, representing classes such as Abelmoschus Esculentus, Acorus Calamus, Aloe Vera, Alstonia Scholaris, and Amaranthus Spinosus. Since the images were sourced from the Google Images search engine, the selected images exhibit a diverse range within each class.

\subsection{Preprocessing Data}
The collected data will undergo preprocessing. Initially, the dataset will be read, followed by subsequent transformations such as resizing, rotating, and normalizing the image data. Images will be resized to dimensions of 128x128 and their pixel values normalized to fall within the range of 0 to 1. Following preprocessing, the data will be split into training and testing sets.

\subsection{Structuring Model}
In this study, we will employ the Transfer Learning method for training. The Transfer Learning models utilized include pre-trained ResNet34 \cite{He2016}, DenseNet121 \cite{Huang2017}, VGG11\_bn \cite{Simonyan2015}, ConvNeXt \cite{convnext}, and Swin Transformer \cite{swin} models. Additionally, we will conduct tests using the scratch model. Let's prepare the code to initiate training using Transfer Learning with these models.

{\color{black}
The architecture of the scratch model begins with an initial convolutional layer that uses 32 output channels and a 3x3 kernel, followed by a 2x2 Max Pooling layer. The second convolutional layer utilizes 64 output channels with a 3x3 kernel, followed by another 2x2 Max Pooling layer. The third convolutional layer employs 128 output channels with a 3x3 kernel, again followed by a 2x2 Max Pooling layer. After the convolutional layers, the output is flattened, transforming the matrix into a vector, which is then passed through a linear classification layer with a ReLU activation function. The final output layer classifies the input into one of 100 classes.}

\subsection{Testing Model}
Testing will be conducted to evaluate the CNN model designed. The testing will comprise both training and testing stages.During the training stage, the CNN model will be evaluated using previously prepared training data. A total of 20,000 and 10,000 images from two different datasets will be utilized. These datasets will be split into 60\% training data and 40\% testing data.After the completion of the training process, testing will commence. For this study, 8,000 and 4,000 images will be used for testing, with 40 images allocated for each class of herbal plants. It's crucial to note that the testing stage will employ different images from those used in the training process to accurately assess the performance of the model.

\section{Experiments}
\subsection{Testing the Vietnam Medicinal Plant Dataset}
In this study, the dataset utilized is the Vietnam Medicinal Plant Dataset public dataset, which will be tested on several models including pre-trained ResNet34 \cite{He2016}, DenseNet121 \cite{Huang2017}, VGG11\_bn \cite{Simonyan2015}, ConvNeXt \cite{convnext}, Swin Transformer \cite{swin}, and the Scratch model.During testing, a dynamic learning rate approach will be employed. The scheduler used for this dynamic learning rate is the ExponentialLR \cite{li2019exponential} scheduler, configured with a gamma value of 0.9. The learning rate for this test is set at 0.001.The dataset being tested consists of a total of 20,000 images, with a training-to-testing data distribution ratio of 60:40, resulting in 12,000 training data and 8,000 testing data. There are 200 classes in this dataset, with each class containing 60 training images and 40 testing images.The testing will span 50 epochs and utilize the Adam optimizer \cite{Kingma2015}, with the Cross Entropy Loss function \cite{zhang2018generalized}.

\begin{table}[H]
\centering
\begin{tabular}{|c|c|l|cc|cc|}
\hline
\multirow{2}{*}{No.} & \multicolumn{1}{c|}{\multirow{2}{*}{Model Name}} & \multicolumn{1}{c|}{\multirow{2}{*}{Resolution}} & \multicolumn{2}{c|}{Training} & \multicolumn{2}{c|}{Testing} \\ \cline{4-7} 
   & \multicolumn{1}{c|}{} &  \multicolumn{1}{c|}{} & \multicolumn{1}{c|}{loss}   & accuracy & \multicolumn{1}{c|}{loss}   & accuracy \\ \hline
   
1. & SIFT \cite{Quoc2021}             & $512^2$&\multicolumn{1}{c|}{-} & -   & 
\multicolumn{1}{c|}{-} & 0.374   \\ \hline
2. & SURF \cite{Quoc2021}             & $512^2$ &\multicolumn{1}{c|}{-} & -   & 
\multicolumn{1}{c|}{-} & 0.347   \\ \hline
3. & SIFT \cite{Quoc2021}             & $256^2$&\multicolumn{1}{c|}{-} & -   & 
\multicolumn{1}{c|}{-} & 0.21   \\ \hline
4. & SURF \cite{Quoc2021}             & $256^2$ &\multicolumn{1}{c|}{-} & -   & 
\multicolumn{1}{c|}{-} & 0.28   \\ \hline
5. & ResNet34              & $128^2$ &\multicolumn{1}{c|}{0.0043} & 0.9995   & 
\multicolumn{1}{c|}{0.8013} & 0.8498   \\ \hline
6. & DenseNet121           &$128^2$ &\multicolumn{1}{c|}{0.0010} & 0.9998   & \multicolumn{1}{c|}{0.5621} & 0.8892   \\ \hline
7. & VGG11\_bn             &$128^2$ &\multicolumn{1}{c|}{0.0245} & 0.9921   & \multicolumn{1}{c|}{1.0136} & 0.8444   \\ \hline
8. & ConvNeXt\_base             & $128^2$ &\multicolumn{1}{c|}{0.0008} & 0.9998   & 
\multicolumn{1}{c|}{0.4098} & 0.9278   \\ \hline
9. & Swin\_t             & $224^2$ &\multicolumn{1}{c|}{0.4905} & 0.8639   & \multicolumn{1}{c|}{1.6539} & 0.6504   \\ \hline
10. & Scratch               &$128^2$ &\multicolumn{1}{c|}{0.7023} & 0.8012   & \multicolumn{1}{c|}{2.7618} & 0.4849   \\ \hline
\end{tabular}
\caption{\textcolor{black}{Training and Testing Results of Vietnam Medicinal Plant Dataset}}
\label{tab:viet_table}
\end{table}

Based on the information presented in Table \ref{tab:viet_table}, which summarizes the performance metrics of various models in our experiment, several key observations can be made regarding their accuracy and architecture. Firstly, when considering all six models in our study, ConvNeXt\_base emerges as the top-performing pre-trained model, boasting an impressive testing accuracy of 92.78\%. This indicates that ConvNeXt\_base excels in accurately classifying images and outperforms the other models under evaluation. Conversely, the model trained from scratch exhibits the lowest testing accuracy among the models, achieving a modest accuracy score of 48.49\%. This outcome can be attributed to the simplicity of the Scratch model's architecture when compared to the more complex, pre-trained models. The Scratch model's limitations in terms of learned features and representations likely contributed to its comparatively lower performance.

Furthermore, when restricting our comparison to the five pre-trained models – ResNet34, VGG11 bn, DenseNet121, ConvNeXt\_base, and Swin\_t where Swin\_t stands out as the model with the lowest accuracy, registering at 65.04\%. While this accuracy rate is respectable, it falls short when contrasted with the superior performance of the other pre-trained models and the difference in image resolution used. It is important to note that the evaluation is not solely based on accuracy but also on other metrics such as training and testing loss. These metrics provide deeper insights into how each model learned and improved during the training and testing phases. For a comprehensive understanding of the model's dynamics, including loss and accuracy graphs, we will delve into a more detailed discussion in the subsequent sections of this paper.

\begin{table}[H]
\centering
\begin{tabular}{|l|l|l|l|l|l|}
\hline
No. & Model Name  & Resolution & Precision & Recall & F1-Score \\ \hline
1.  & ResNet34    & $128^2$        & 0.8515    & 0.8281 & 0.8281  \\ \hline
2.  & DenseNet121 & $128^2$        & 0.9141    & 0.9063 & 0.9063   \\ \hline
3.  & VGG11\_bn   & $128^2$        & 0.9218    & 0.8750 & 0.8854   \\ \hline
4.  & ConvNeXt    & $128^2$        & 0.9766    & 0.9531 & 0.9583   \\ \hline
5.  & Swin\_t     & $224^2$        & 0.7083    & 0.7188 & 0.7073   \\ \hline
6.  & Scratch     & $128^2$        & 0.4766    & 0.4688 & 0.4635   \\ \hline
\end{tabular}
\caption{\textcolor{black}{Evaluation Metrics of Vietnam Medicinal Plant Dataset}}
\label{tab:viet_table_eval}
\end{table}

{\color{black}
Based on Table \ref{tab:viet_table_eval}, the presented results offer a comprehensive evaluation of diverse neural network models in the context of a specific task, utilizing key performance metrics such as precision, recall, and F1-Score. The models under consideration encompass well-established architectures, including ResNet34, DenseNet121, VGG11 bn, ConvNeXt, and Swin t, along with a model trained from scratch denoted as "Scratch."

Given the reported metrics of Precision, Recall, and F1-Score, along with the corresponding model names and resolutions. In the first model, ResNet34, with a resolution of $128^2$, achieves a well-balanced performance, demonstrating competitive precision, recall, and F1-Score. The reported values suggest that ResNet34 is effective in identifying and classifying instances, with a slightly higher focus on precision.

Moving on to the second model, DenseNet121, also with a resolution of 1282, stands out with high precision, recall, and F1-Score. These results indicate that DenseNet121 excels in capturing both positive and negative instances, with a particular emphasis on precision. The resolution of $128^2$ signifies the input size used for this model in the evaluation.

The third model, VGG11\_bn, at a resolution of $128^2$, demonstrates notable precision but slightly lower recall and F1-Score. While its precision is high, the trade-off appears to be a compromise in recall. The values suggest that VGG11\_bn is effective in minimizing false positives, but it may miss some instances, impacting its recall and overall F1-Score.

The fourth model, ConvNeXt, with a resolution of $128^2$, outperforms others with impressive precision, recall, and F1-Score. These results indicate that ConvNeXt excels in both identifying and capturing instances with high accuracy, making it a robust choice for the given task.

However, the fifth model, Swin\_t, with a higher resolution of $224^2$, exhibits lower precision, recall, and F1-Score values. This suggests that increasing the resolution may not have positively impacted the model's performance in this context, potentially indicating a trade-off between resolution and classification accuracy.

Finally, the sixth model, Scratch, at a resolution of $128^2$, displays comparatively lower precision, recall, and F1-Score values. These results indicate that the Scratch model, developed without leveraging pre-trained weights, may not perform as well as the other models evaluated in this study.}

\subsubsection{Testing Vietnam Dataset with pre-trained model ResNet34}

From the tests conducted using the pre-trained ResNet34 model, the accuracy level of the testing data was found to be 85.58\%. The graph depicting the loss and accuracy values is shown in the following figure.

\begin{figure}[H]
    \centering
    \includegraphics[width=7cm]{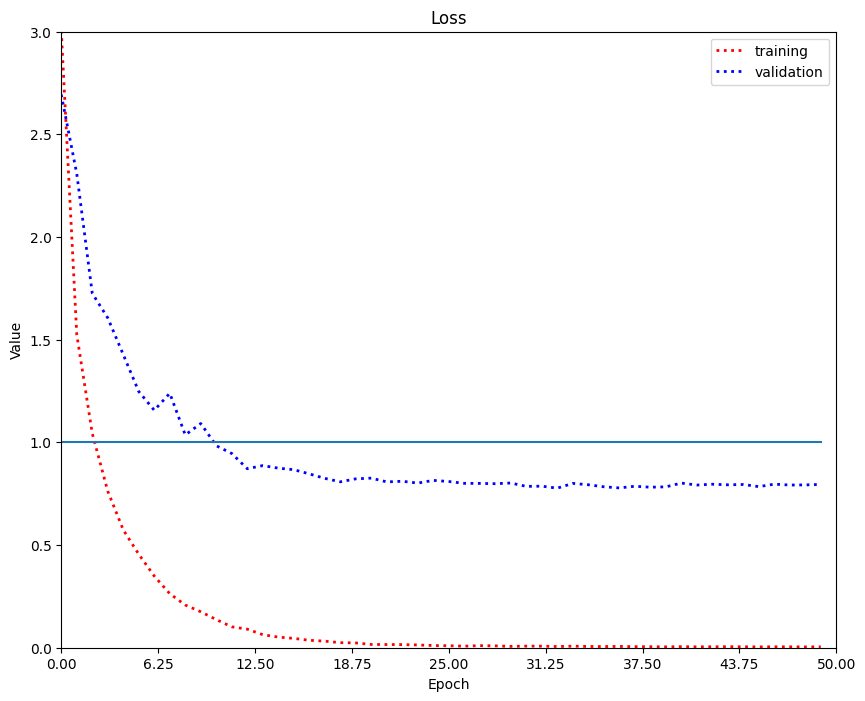}
    \caption{Graph of ResNet34 Model Loss Value from Vietnam Dataset Against Epoch}
    \label{viet_resnet_loss}
\end{figure}

Based on Figure \ref{viet_resnet_loss}, it is evident that as the number of epochs increases, there is a consistent decrease in the loss value during both the training and testing processes. This observation indicates that the training process effectively minimizes the loss value, thereby mitigating overfitting.

\begin{figure}[H]
    \centering
    \includegraphics[width=7cm]{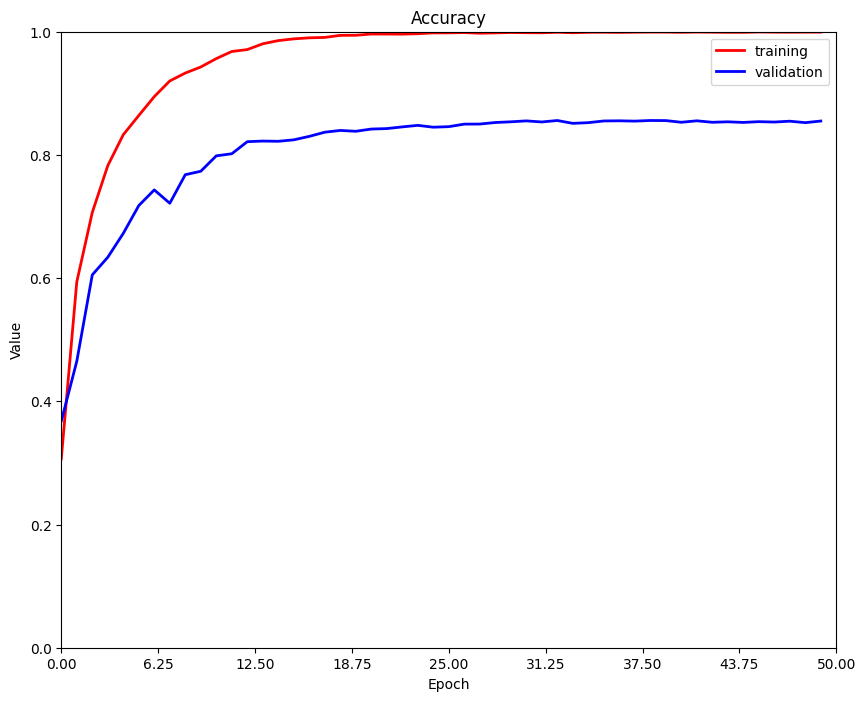}
    \caption{Graph of ResNet34 Model Accuracy Value from Vietnam Dataset Against Epoch}
    \label{viet_resnet_acc}
\end{figure}

Based on Figure \ref{viet_resnet_acc}, it's evident that the model achieves a relatively high accuracy as the number of epochs increases. The accuracy of the training data surpasses 0.9 when the epoch exceeds 12 and continues to increase with each subsequent epoch. Similarly, the testing data attains an accuracy level above 0.8 when the epoch surpasses 15. This upward trend in accuracy values suggests that the training process is progressing effectively.

\subsubsection{Testing Vietnam Dataset with pre-trained model DenseNet121}
The tests conducted using the pre-trained DenseNet121 model yielded an accuracy level of 88.69\% on the testing data. Below is the graph illustrating the loss and accuracy values.

\begin{figure}[H]
    \centering
    \includegraphics[width=7cm]{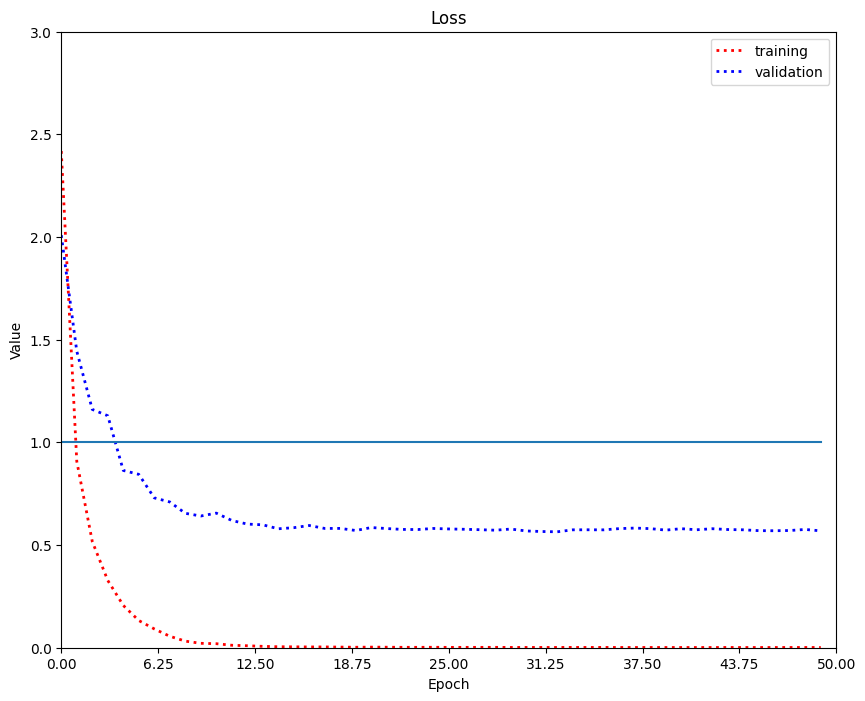}
    \caption{Graph of DenseNet121 Model Loss Value from Vietnam Dataset Against Epoch}
    \label{viet_densenet_loss}
\end{figure}

Based on Figure \ref{viet_densenet_loss}, it is evident that the loss value decreases consistently throughout both the training and testing processes. This indicates that the training process effectively minimizes the loss value, suggesting a successful training procedure.

\begin{figure}[H]
    \centering
    \includegraphics[width=7cm]{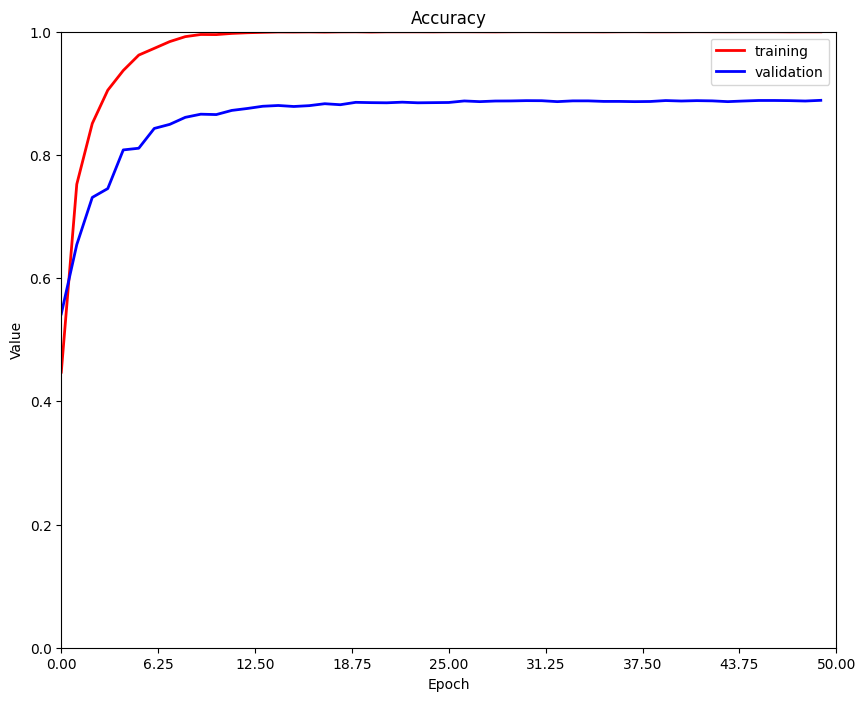}
    \caption{Graph of DenseNet121 Model Accuracy Value from Vietnam Dataset Against Epoch}
    \label{viet_densenet_acc}
\end{figure}

Based on Figure \ref{viet_densenet_acc}, it's apparent that the model achieves high accuracy, surpassing 0.9 on the training data by the 5th epoch. Similarly, on the testing data, it achieves an accuracy level above 0.8 when the epoch surpasses 7. This consistent increase in accuracy values indicates that the training process is progressing effectively.

\subsubsection{Testing Vietnam Dataset with pre-trained model VGG\_bn}
The tests conducted using the VGG11\_bn pre-trained model resulted in an accuracy level of 84.53\% on the testing data. Below is the graph illustrating the loss and accuracy of the model across epochs.

\begin{figure}[H]
    \centering
    \includegraphics[width=7cm]{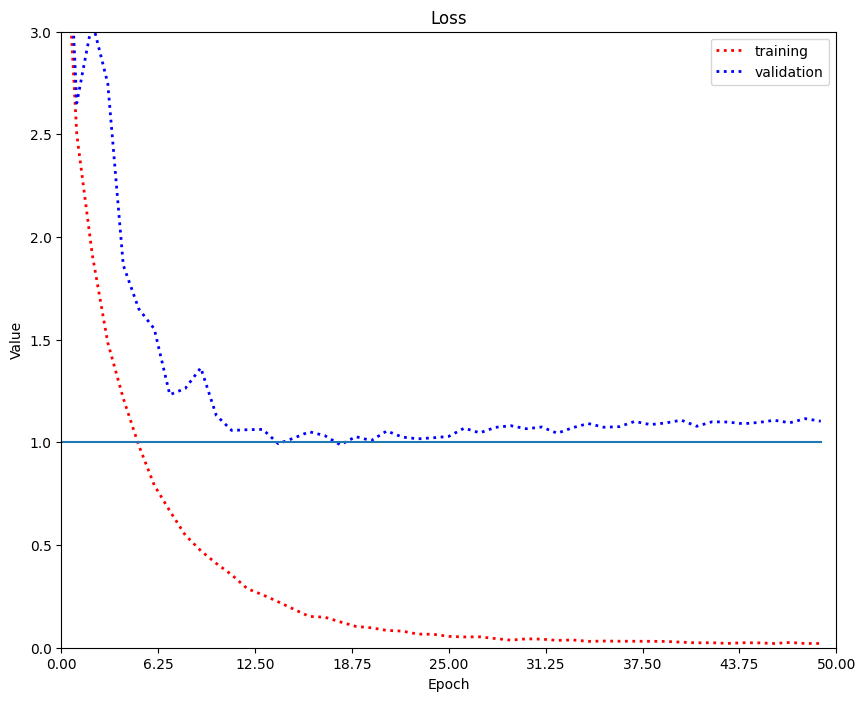}
    \caption{Graph of VGG11\_bn Model Loss Value from Vietnam Dataset Against Epoch}
    \label{viet_vgg_loss}
\end{figure}

Based on Figure \ref{viet_vgg_loss}, it is apparent that an increasing number of epochs result in a decreasing loss value during both the training and testing processes. However, compared to the two previous pre-trained models, the loss value in the VGG11\_bn pre-trained model appears to be higher. This suggests that while the training process effectively minimizes loss, it may not perform as optimally as the other pre-trained models.

\begin{figure}[H]
    \centering
    \includegraphics[width=7cm]{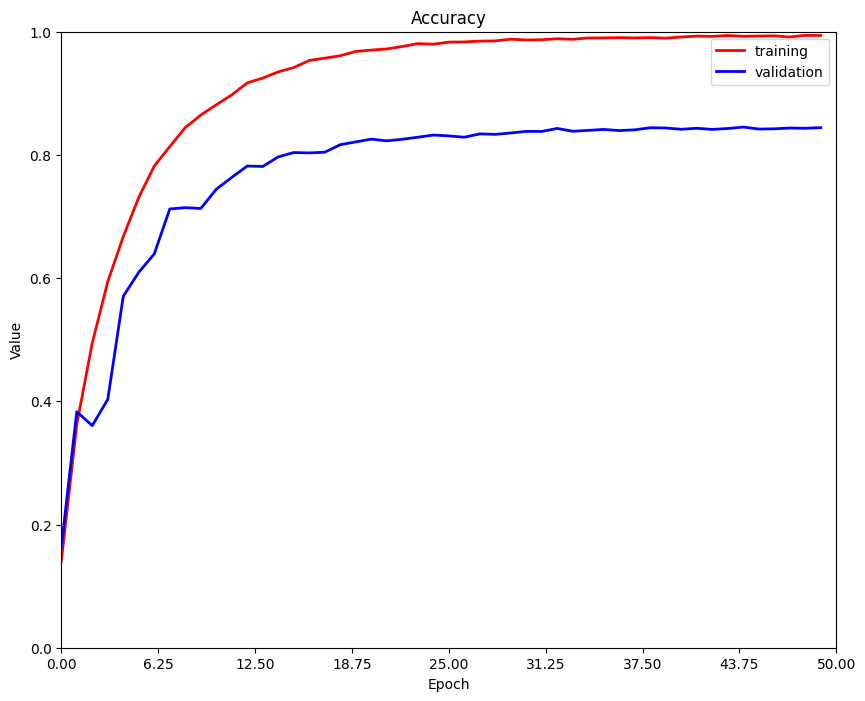}
    \caption{Graph of VGG11\_bn Model Accuracy Value from Vietnam Dataset Against Epoch}
    \label{viet_vgg_acc}
\end{figure}

Based on Figure \ref{viet_vgg_acc}, it's evident that the model achieves high accuracy as the number of epochs increases. The accuracy of the training data surpasses 0.9 when the epoch exceeds around 15 and continues to increase thereafter. However, in the testing data, it only achieves an accuracy level above 0.8 when the epoch surpasses 18. Despite the increase in accuracy value indicating a successful training process, it requires more epochs to achieve a high score compared to other pre-trained models.

{\color{black}
\subsubsection{Testing Vietnam Dataset with pre-trained model ConvNeXt\_base}

The tests conducted using the ConvNeXt\_base pre-trained model yielded an accuracy level of 92.84\% on the testing data. Below is the graph illustrating the loss and accuracy values.}

\begin{figure}[H]
    \centering
    \includegraphics[width=7cm]{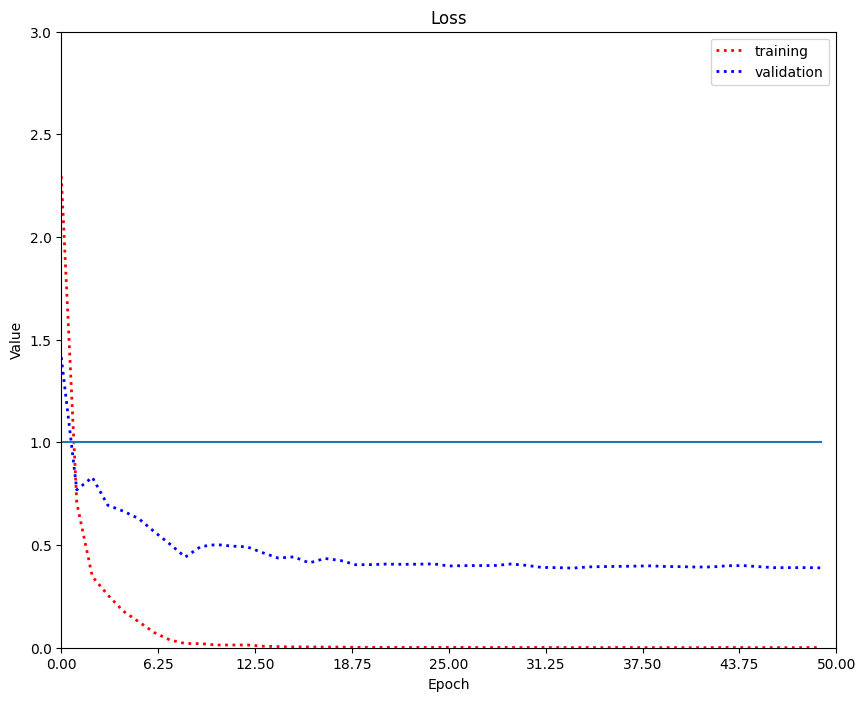}
    \caption{\color{black}{Graph of ConvNeXt\_base Model Loss Value from Vietnam Dataset Against Epoch}}
    \label{viet_convnext_loss}
\end{figure}

{\color{black}
Based on Figure \ref{viet_convnext_loss}, it's evident that an increasing number of epochs results in a decreasing loss value during both the training and testing processes. This observation suggests that the training process effectively minimizes the loss value, indicating successful model training.}

\begin{figure}[H]
    \centering
    \includegraphics[width=7cm]{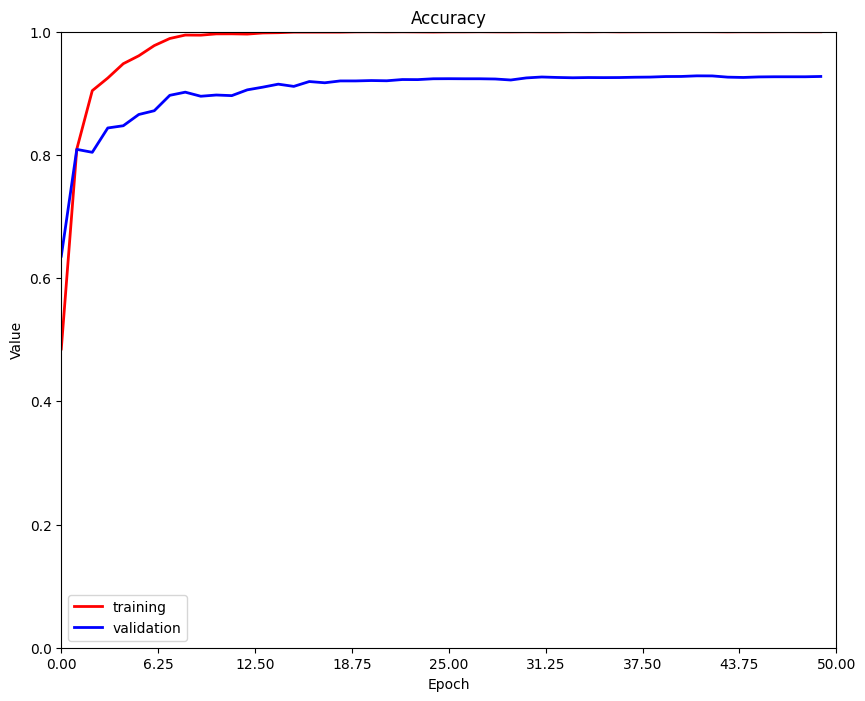}
    \caption{\color{black}{Graph of ConvNeXt\_base Model Accuracy Value from Vietnam Dataset Against Epoch}}
    \label{viet_convnext_acc}
\end{figure}

{\color{black}
Based on Figure \ref{viet_convnext_acc}, it's apparent that the model achieves a fairly high accuracy as the number of epochs increases. The accuracy of the training data surpasses 0.9 when the epoch exceeds 4 and continues to increase thereafter. Similarly, in the testing data, it achieves an accuracy level above 0.9 when the epoch surpasses 13. This increase in accuracy values indicates that the training process is progressing effectively.}

{\color{black}
\subsubsection{Testing Vietnam Dataset with pre-trained model Swin\_t}
The tests conducted using the Swin\_t pre-trained model resulted in an accuracy level of 79.01\% on the testing data. Below is the graph illustrating the loss and accuracy values.}

\begin{figure}[H]
    \centering
    \includegraphics[width=7cm]{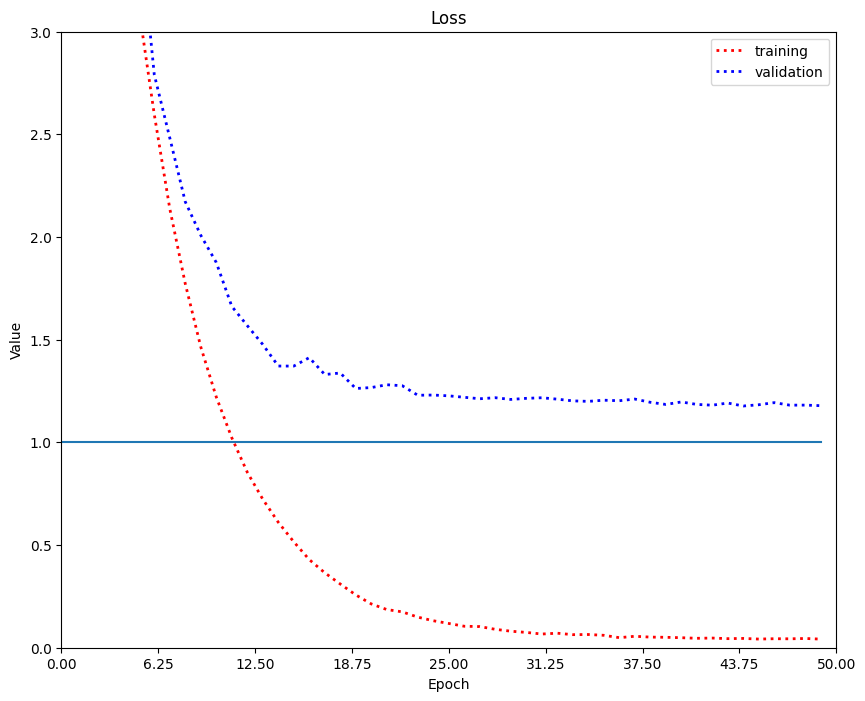}
    \caption{\color{black}{Graph of Swin\_t Model Loss Value from Vietnam Dataset Against Epoch}}
    \label{viet_swin_loss}
\end{figure}

{\color{black}
Based on Figure \ref{viet_swin_loss}, it's evident that the loss value decreases during both the training and testing processes, although the testing data process still exhibits relatively high loss values above 1 even after 50 epochs.}

\begin{figure}[H]
    \centering
    \includegraphics[width=7cm]{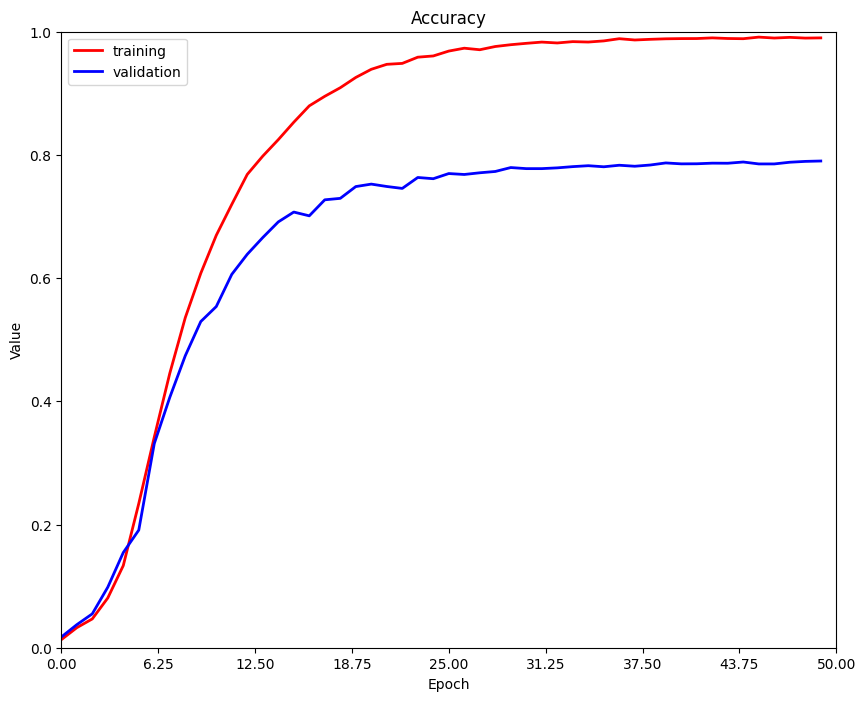}
    \caption{\color{black}{Graph of Swin\_t Model Accuracy Value from Vietnam Dataset Against Epoch}}
    \label{viet_swin_acc}
\end{figure}

{\color{black}
Based on Figure \ref{viet_swin_acc}, it's apparent that the model achieves high accuracy, surpassing 0.9 starting from the 20th epoch for the training data. However, in the testing data, it cannot achieve an accuracy value of more than 0.8 until the end of the epoch.}

\subsubsection{Testing Vietnam Dataset with Scratch model}
The tests conducted using the Scratch model revealed a relatively low accuracy rate of 37.63\% on the testing data. Below is the graph illustrating the loss and accuracy of the model across epochs.

\begin{figure}[H]
    \centering
    \includegraphics[width=7cm]{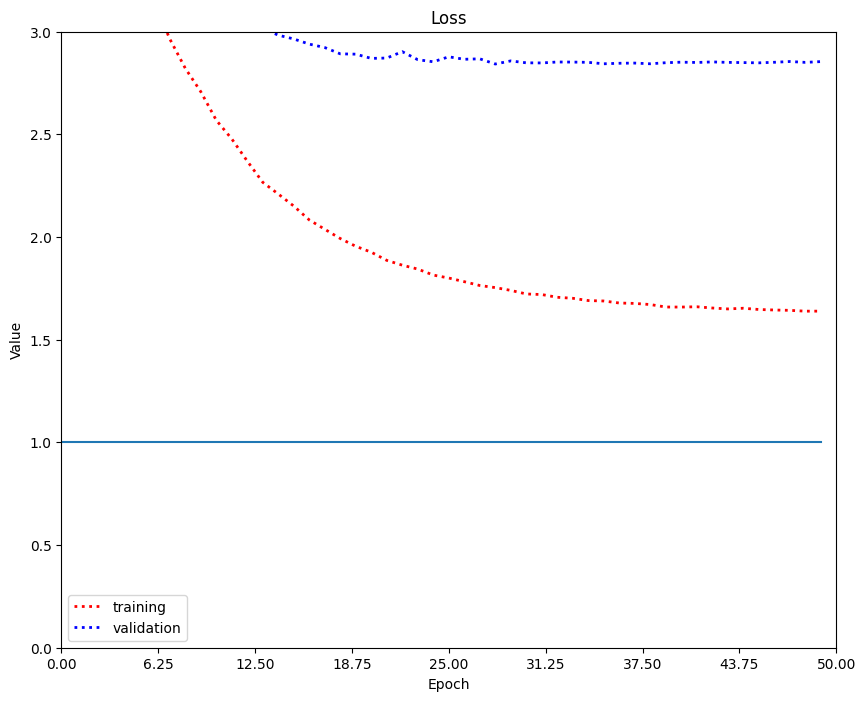}
    \caption{Graph of Scratch Model Loss Value from Vietnam Dataset Against Epoch}
    \label{viet_scratch_loss}
\end{figure}

Based on Figure \ref{viet_scratch_loss}, it's evident that the loss value of the scratch model is notably high. In comparison to the pre-trained models tested, the loss value in this scratch model exhibits a much wider range, indicating a significantly higher loss. This discrepancy is attributed to the simplicity of the scratch model compared to the pre-trained models. Achieving a smaller loss value typically requires a more complex and pre-trained model.

\begin{figure}[H]
    \centering
    \includegraphics[width=7cm]{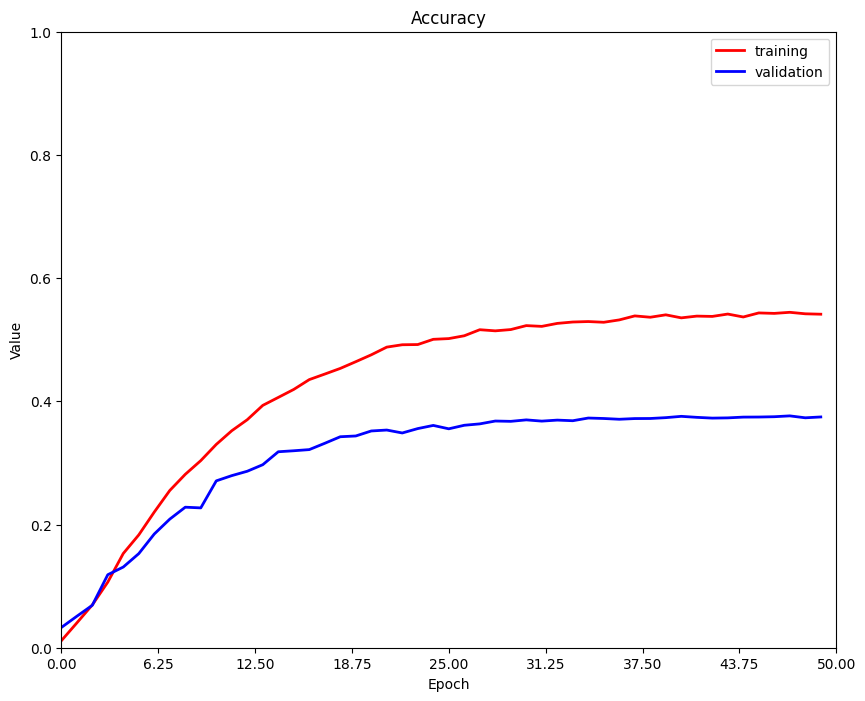}
    \caption{Graph of Scratch Model Accuracy Value from Vietnam Dataset Against Epoch}
    \label{viet_scratch_acc}
\end{figure}

Based on Figure \ref{viet_scratch_acc}, it's apparent that the accuracy value of the scratch model is relatively low. Similar to the loss value, the accuracy values of this scratch model exhibit a wide range compared to the accuracy values of the other tested pre-trained models. This emphasizes the need for a more complex model, such as a pre-trained model, to achieve higher accuracy values.

\subsection{Testing the Indonesia Medicinal Plant Dataset}
In this research, a self-collected dataset will be tested on the same models as the Vietnam Medicinal Plant dataset, including pre-trained ResNet34 \cite{He2016}, DenseNet121 \cite{Huang2017}, VGG11\_bn \cite{Simonyan2015}, ConvNeXt \cite{convnext}, Swin Transformer \cite{swin}, and the Scratch model. Dynamic learning rate will also be employed in this test, utilizing the ExponentialLR \cite{li2019exponential} scheduler with a gamma configuration of 0.9. The learning rate for this test remains 0.001.The self-collected dataset comprises 10,000 images, with a training-to-testing data distribution ratio of 60:40, resulting in 6,000 training data and 4,000 testing data. There are 100 classes in this dataset. This test will span 50 epochs, utilizing the Adam optimizer \cite{Kingma2015}, with the Cross Entropy Loss function \cite{zhang2018generalized}.

\begin{table}[H]
\centering
\begin{tabular}{|c|c|l|cc|cc|}
\hline
\multirow{2}{*}{No.} & \multicolumn{1}{c|}{\multirow{2}{*}{Model Name}} & \multicolumn{1}{c|}{\multirow{2}{*}{Resolution}} & \multicolumn{2}{c|}{Training} & \multicolumn{2}{c|}{Testing} \\ \cline{4-7} 
   & \multicolumn{1}{c|}{} &  \multicolumn{1}{c|}{} & \multicolumn{1}{c|}{loss}   & accuracy & \multicolumn{1}{c|}{loss}   & accuracy \\ \hline
1. & ResNet34              & $128^2$ &\multicolumn{1}{c|}{0.0143} & 0.9965   & \multicolumn{1}{c|}{0.6857} & 0.8650   \\ \hline
2. & DenseNet121           & $128^2$ &\multicolumn{1}{c|}{0.0027} & 0.9998   & \multicolumn{1}{c|}{0.4873} & 0.8910   \\ \hline
3. & VGG11\_bn             & $128^2$ & \multicolumn{1}{c|}{0.0301} & 0.9898   & \multicolumn{1}{c|}{0.8412} & 0.8703   \\ \hline
4. & ConvNeXt\_Base             & $128^2$ &\multicolumn{1}{c|}{0.0026} & 0.9995   &  \multicolumn{1}{c|}{0.3622} & 0.9250   \\ \hline
5. & Swin\_t             & $224^2$ &\multicolumn{1}{c|}{0.1705} & 0.9562   & \multicolumn{1}{c|}{1.1918} & 0.7655   \\ \hline
6. & Scratch               & $128^2$ &\multicolumn{1}{c|}{0.7147} & 0.8162   & \multicolumn{1}{c|}{2.3606} & 0.5390   \\ \hline
\end{tabular}
\caption{\color{black}{Training and Testing Results of Indonesia Medicinal Plant Dataset}}
\label{tab:indo_table}
\end{table}

Based on Table \ref{tab:indo_table}, it can be seen from all the models that have been tested, the ConvNeXt\_base model with an accuracy of 92.5\% in the testing process is the model with the highest accuracy compared to the other models. Just like in testing the Vietnam Medicinal Plant Dataset, the Scratch model is the model with the lowest accuracy compared to the other models. Then of the five pre-trained models, the Swin\_t model is the model that has the lowest accuracy, namely 76.55\%. The loss and accuracy graphs of each model can be seen in the following discussion.

Furthermore, when restricting our comparison to the five pre-trained models – ResNet34, VGG11 bn, DenseNet121, ConvNeXt\_base, and Swin\_t – Swin\_t stands out as the model with the lowest accuracy, registering at 76.55\%. For a comprehensive understanding of the model's dynamics, including loss and accuracy graphs, we will delve into a more detailed discussion in the subsequent sections of this paper.

\begin{table}[H]
\centering
\begin{tabular}{|l|l|l|l|l|l|}
\hline
No. & Model Name  & Resolution & Precision & Recall & F1-Score \\ \hline
1.  & ResNet34    & $128^2$        & 0.8995    & 0.8750 & 0.8730   \\ \hline
2.  & DenseNet121 & $128^2$        & 0.9531    & 0.9188 & 0.9241   \\ \hline
3.  & VGG11\_bn   & $128^2$        & 0.8828    & 0.8375 & 0.8428   \\ \hline
4.  & ConvNeXt    & $128^2$        & 0.9594    & 0.9375 & 0.9434   \\ \hline
5.  & Swin\_t     & $224^2$        & 0.8849    & 0.8188 & 0.8181   \\ \hline
6.  & Scratch     & $128^2$        & 0.6536    & 0.5125 & 0.5420   \\ \hline
\end{tabular}
\caption{\color{black}{Evaluation Metrics of Indonesia Medicinal Plant Dataset}}
\label{tab:indo_table_eval}
\end{table}

{\color{black}
Based on Table \ref{tab:indo_table_eval}, the presented findings constitute an in-depth evaluation of multiple neural network models within the scope of a specific task, with a particular emphasis on precision, recall, and F1-Score. Each model, including ResNet34, DenseNet121, VGG11 bn, ConvNeXt, Swin t, and a Scratch model trained from scratch, was rigorously assessed for its efficacy in positive predictions and overall performance.

ResNet34 demonstrated a precision of 0.8995, indicating a strong accuracy in positive predictions, and a recall of 0.8750, showcasing its ability to capture around 87.50\% of the actual positive instances. The corresponding F1-Score of 0.8730 reflects a well-balanced trade-off between precision and recall, essential for comprehensive model assessment. DenseNet121 exhibited a higher precision of 0.9531, signifying an enhanced accuracy in positive predictions compared to ResNet34. The recall of 0.9188 highlights the model's proficiency in capturing approximately 91.88\% of the actual positive instances. The resultant F1-Score of 0.9241 further emphasizes the model's balanced performance. VGG11 bn, although displaying a precision of 0.8828, indicating a commendable accuracy in positive predictions, showed a slightly lower recall of 0.8375. The resultant F1-Score of 0.8428 underscores the model's ability to strike a balance between precision and recall.

ConvNeXt, with a precision of 0.9594, demonstrated a high accuracy in positive predictions, and a recall of 0.9375, indicating a commendable capacity to capture 93.75\% of the actual positive instances. The F1-Score of 0.9434 corroborates ConvNeXt's robust overall performance. Swin t, with a precision of 0.8849, a recall of 0.8188, and an F1-Score of 0.8181, suggests a noteworthy accuracy in positive predictions, but a comparatively lower capacity to capture the entirety of actual positive instances. The Scratch model, trained from scratch, exhibited a precision of 0.6536, a recall of 0.5125, and an F1-Score of 0.5420. These results indicate a significant discrepancy, suggesting challenges in accurately identifying positive instances, potentially stemming from the training process or model architecture.}

\subsubsection{Testing Indonesia Dataset with pre-trained model ResNet34}
The tests conducted using the pre-trained ResNet34 model yielded an accuracy level of 85.65\% on the testing data. Below is the graph illustrating the loss and accuracy values.

\begin{figure}[H]
    \centering
    \includegraphics[width=7cm]{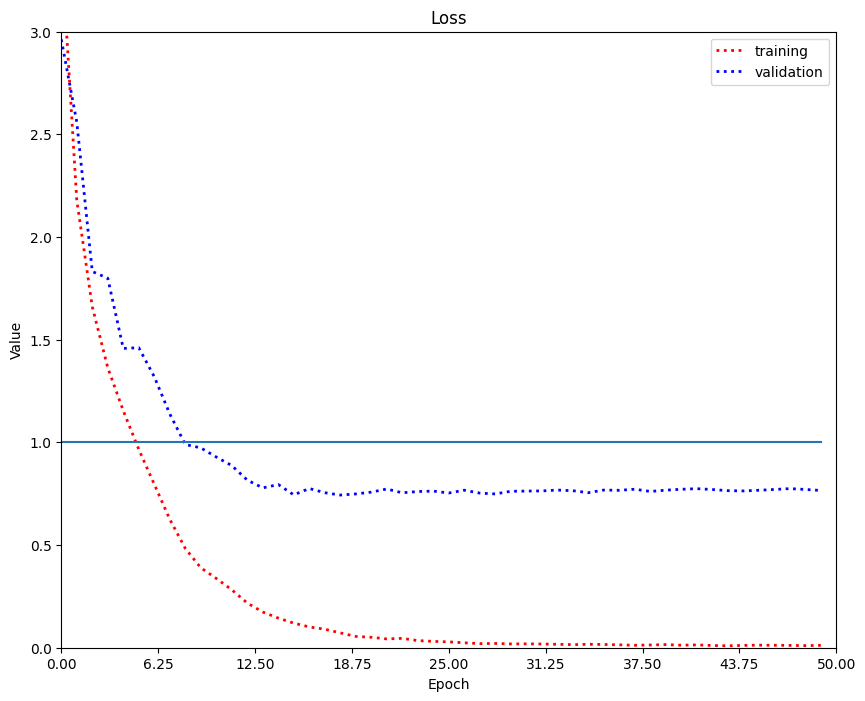}
    \caption{Graph of ResNet34 Model Loss Value from Indonesia Dataset Against Epoch}
    \label{indo_resnet_loss}
\end{figure}

Based on Figure \ref{indo_resnet_loss}, it's evident that as the number of epochs increases, the loss value decreases during both the training and testing processes. The training process appears to be more effective in minimizing loss values compared to the testing process, indicating successful model training.

\begin{figure}[H]
    \centering
    \includegraphics[width=7cm]{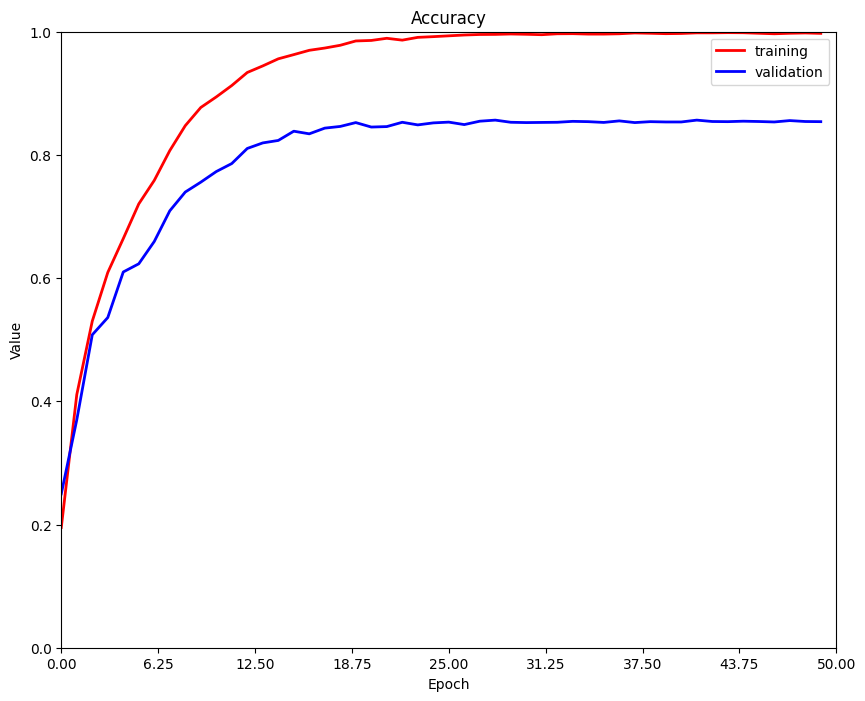}
    \caption{Graph of ResNet34 Model Accuracy Value from Indonesia Dataset Against Epoch}
    \label{indo_resnet_acc}
\end{figure}

Based on Figure \ref{indo_resnet_acc}, it's apparent that the model achieves higher accuracy with an increasing number of epochs. The accuracy value of the training data begins to exceed 0.9 around the 13th epoch, indicating effective model training. Similarly, the testing data achieves an accuracy level of 0.8 when the epoch surpasses 12. This increase in accuracy values suggests that the training process is progressing effectively.

\subsubsection{Testing Indonesia Dataset with pre-trained model DenseNet121}
The tests conducted using the pre-trained DenseNet121 model resulted in an accuracy rate of 87.4\% on the testing data. Below is the graph illustrating the loss and accuracy values.

\begin{figure}[H]
    \centering
    \includegraphics[width=7cm]{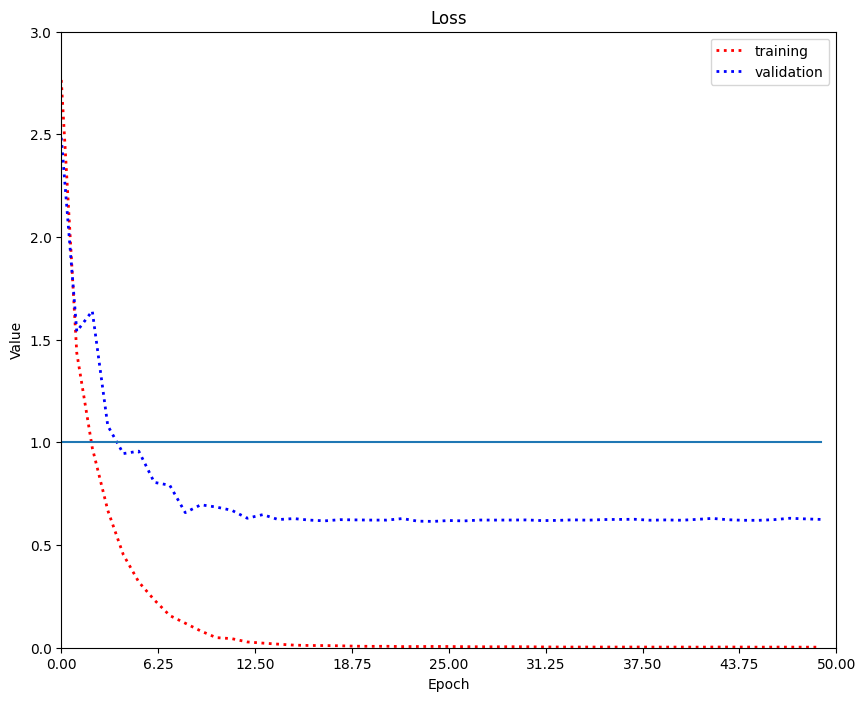}
    \caption{Graph of DenseNet121 Model Loss Value from Indonesia Dataset Against Epoch}
    \label{indo_densenet_loss}
\end{figure}

Based on Figure \ref{indo_densenet_loss}, it's evident that the loss value decreases steadily during the training process. However, in testing, the loss value begins to plateau when it reaches a value of 0.6. This indicates that the training process on DenseNet121 effectively minimizes the loss value, although the testing process struggles to decrease further once it reaches 0.6.

\begin{figure}[H]
    \centering
    \includegraphics[width=7cm]{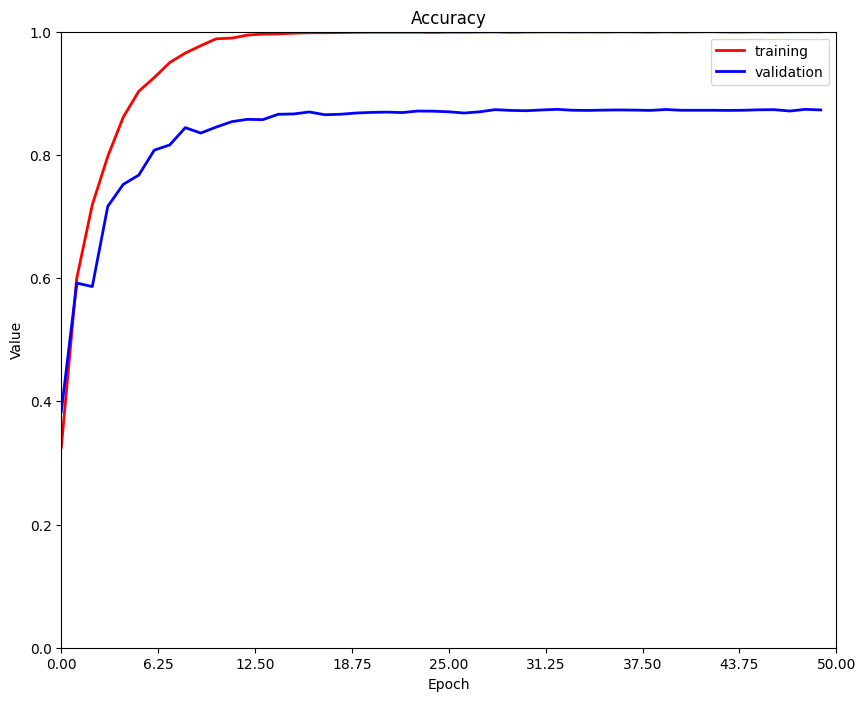}
    \caption{Graph of DenseNet121 Model Accuracy Value from Indonesia Dataset Against Epoch}
    \label{indo_densenet_acc}
\end{figure}

Based on Figure \ref{indo_densenet_acc}, it's evident that the training process consistently achieves high accuracy, surpassing 0.9 when the epoch exceeds 7. However, in the testing data, it only achieves an accuracy level above 0.8 when the epoch surpasses 6. Despite this, the increase in accuracy values indicates that the training process progressed effectively.

\subsubsection{Testing Indonesia Dataset with pre-trained model VGG11\_bn}
The tests conducted using the pre-trained VGG11\_bn model resulted in an accuracy of 82\% on the testing data. Below is the graph illustrating the loss and accuracy of the model across epochs.

\begin{figure}[H]
    \centering
    \includegraphics[width=7cm]{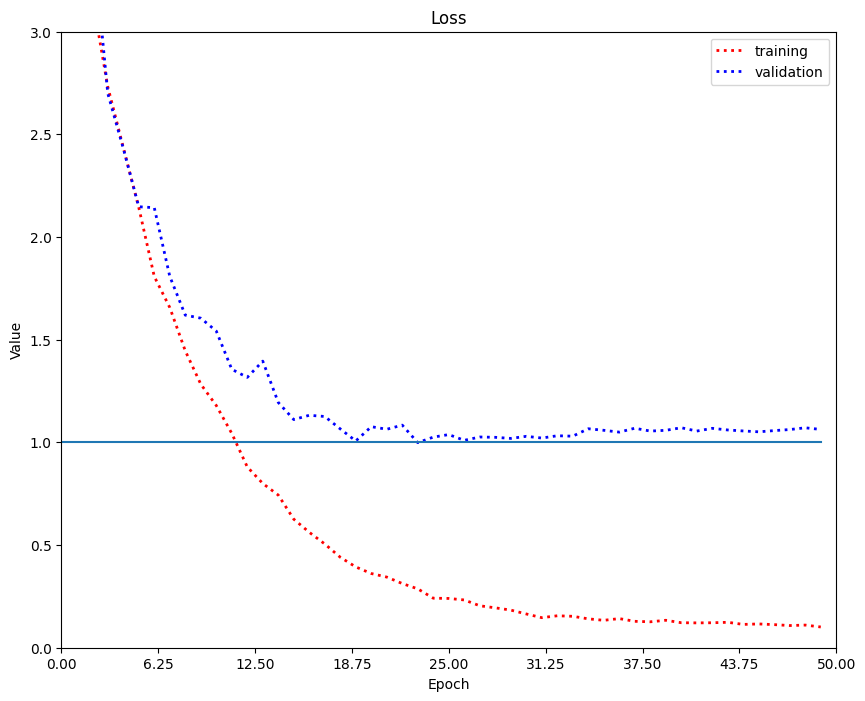}
    \caption{Graph of VGG11\_bn Model Loss Value from Indonesia Dataset Against Epoch}
    \label{indo_vgg_loss}
\end{figure}

Based on Figure \ref{indo_vgg_loss}, it's evident that the loss value decreases as the epoch increases during both the training and testing processes. However, compared to other pre-trained models, the loss value in the VGG pre-trained model appears to be higher. This suggests that while the training process effectively minimizes loss values, it may not perform as optimally as other pre-trained models in minimizing loss values.

\begin{figure}[H]
    \centering
    \includegraphics[width=7cm]{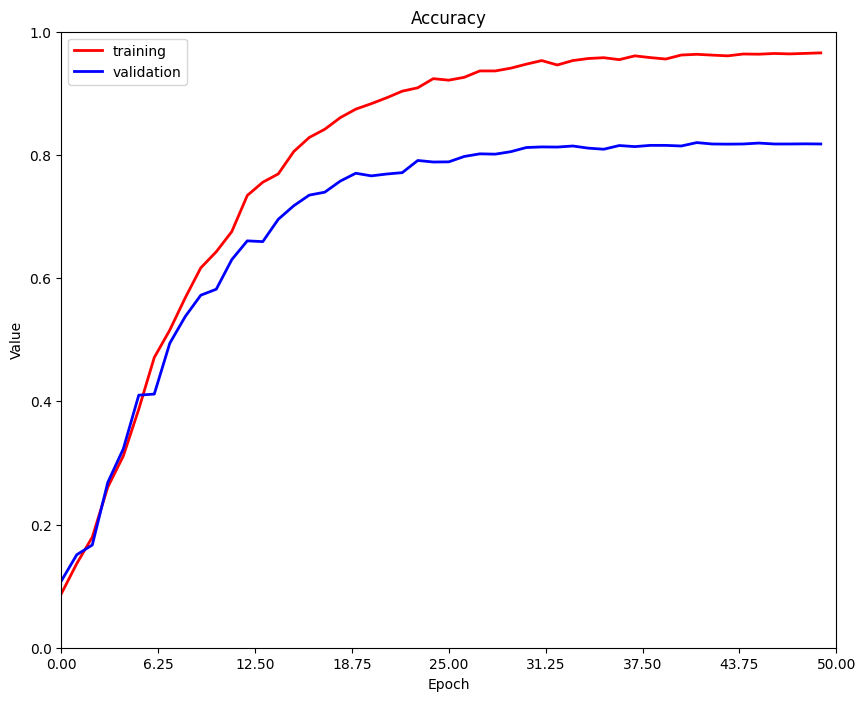}
    \caption{Graph of VGG11\_bn Model Accuracy Value from Indonesia Dataset Against Epoch}
    \label{indo_vgg_acc}
\end{figure}

Based on Figure \ref{indo_vgg_acc}, it's evident that the model achieves higher accuracy as the number of epochs increases. The accuracy value of the training data exceeds 0.9 when the epoch surpasses around 17. However, in the testing data, it only begins to reach an accuracy level of 0.8 consistently when the epoch exceeds 17. This highlights the necessity for an increasing number of epochs to attain high accuracy values.

{\color{black}
\subsubsection{Testing Indonesia Dataset with pre-trained model ConvNeXt\_base}

The tests conducted using the pre-trained ConvNeXt\_base model resulted in an accuracy level of 91.23\% on the testing data. Below is the graph illustrating the loss and accuracy values.}

\begin{figure}[H]
    \centering
    \includegraphics[width=7cm]{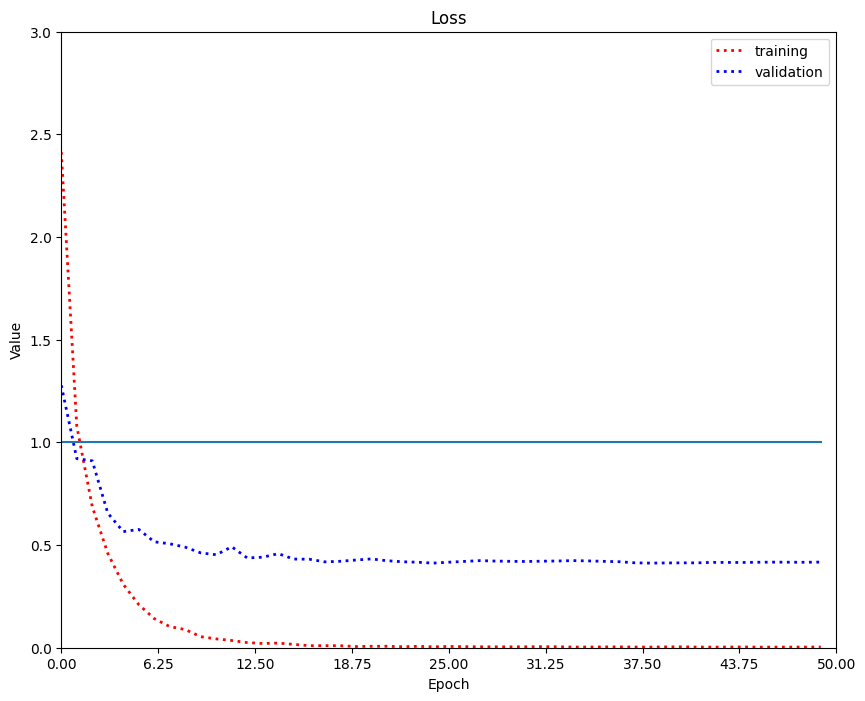}
    \caption{\color{black}{Graph of ConvNeXt\_base Model Loss Value from Indonesia Dataset Against Epoch}}
    \label{indo_convnext_loss}
\end{figure}

{\color{black}
Based on Figure \ref{indo_convnext_loss}, it's evident that as the number of epochs increases, the loss value decreases during both the training and testing processes. This observation suggests that the training process effectively minimizes the loss value, indicating successful model training.}

\begin{figure}[H]
    \centering
    \includegraphics[width=7cm]{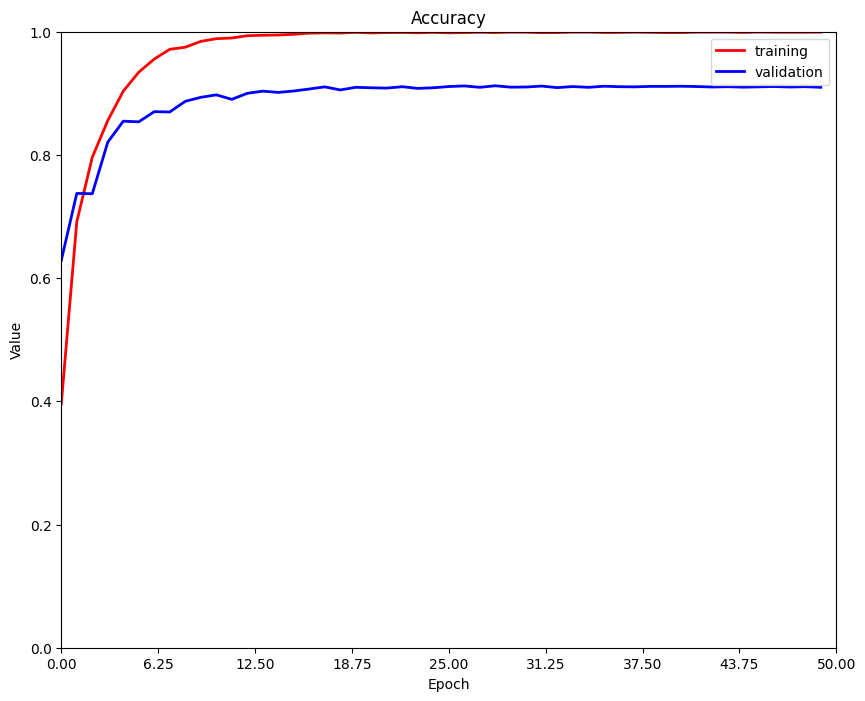}
    \caption{\color{black}{Graph of ConvNeXt\_base Model Accuracy Value from Indonesia Dataset Against Epoch}}
    \label{indo_convnext_acc}
\end{figure}

{\color{black}
Based on Figure \ref{indo_convnext_acc}, it's apparent that the model achieves a fairly high accuracy as the number of epochs increases. The accuracy value of the training data surpasses 0.9 when the epoch exceeds 5 and continues to increase thereafter. Similarly, in the testing data, it achieves an accuracy value above 0.9 when the epoch surpasses 15. This increase in accuracy values indicates that the training process is progressing effectively.

\subsubsection{Testing Indonesia Dataset with pre-trained model Swin\_t}
The tests conducted using the pre-trained Swin\_t model resulted in an accuracy level of 66.95\% on the testing data. Below is the graph illustrating the loss and accuracy values.}

\begin{figure}[H]
    \centering
    \includegraphics[width=7cm]{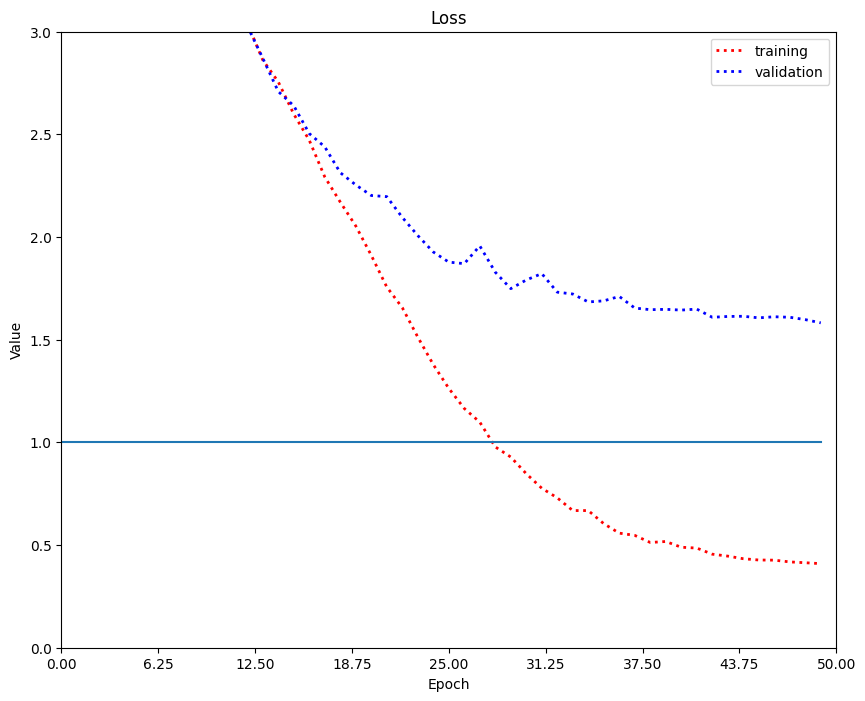}
    \caption{\color{black}{Graph of Swin\_t Model Loss Value from Indonesia Dataset Against Epoch}}
    \label{indo_swin_loss}
\end{figure}

{\color{black}
Based on Figure \ref{indo_swin_loss}, it's evident that the loss value decreases during both the training and testing processes, although the testing data process still exhibits relatively high loss values above 1 even after 50 epochs.}

\begin{figure}[H]
    \centering
    \includegraphics[width=7cm]{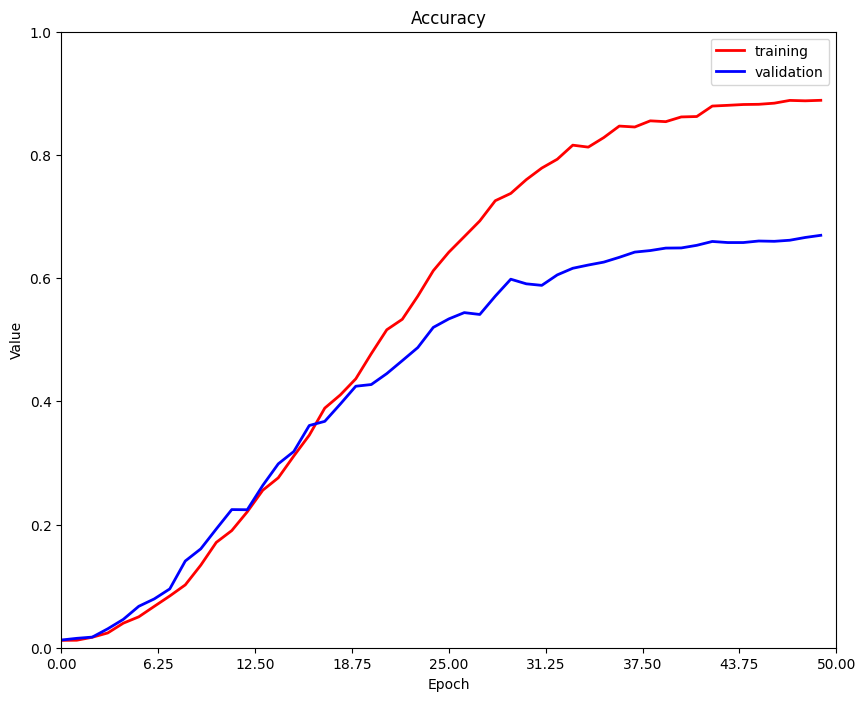}
    \caption{\color{black}{Graph of Swin\_t Model Accuracy Value from Indonesia Dataset Against Epoch}}
    \label{indo_swin_acc}
\end{figure}

{\color{black}
Based on Figure \ref{indo_swin_acc}, it's apparent that the model achieves high accuracy, surpassing 0.8 starting from the 34th epoch for the training data. However, in the testing data, it cannot achieve an accuracy value of more than 0.7 until the end of the epoch.}

\subsubsection{Testing Indonesia Dataset with Scratch model}
The tests conducted using the Scratch model on the Indonesia Medicinal Plant Dataset resulted in a small accuracy level of 43.53\% on the testing data. Below is the graph illustrating the loss and accuracy of the model across epochs.

\begin{figure}[H]
    \centering
    \includegraphics[width=7cm]{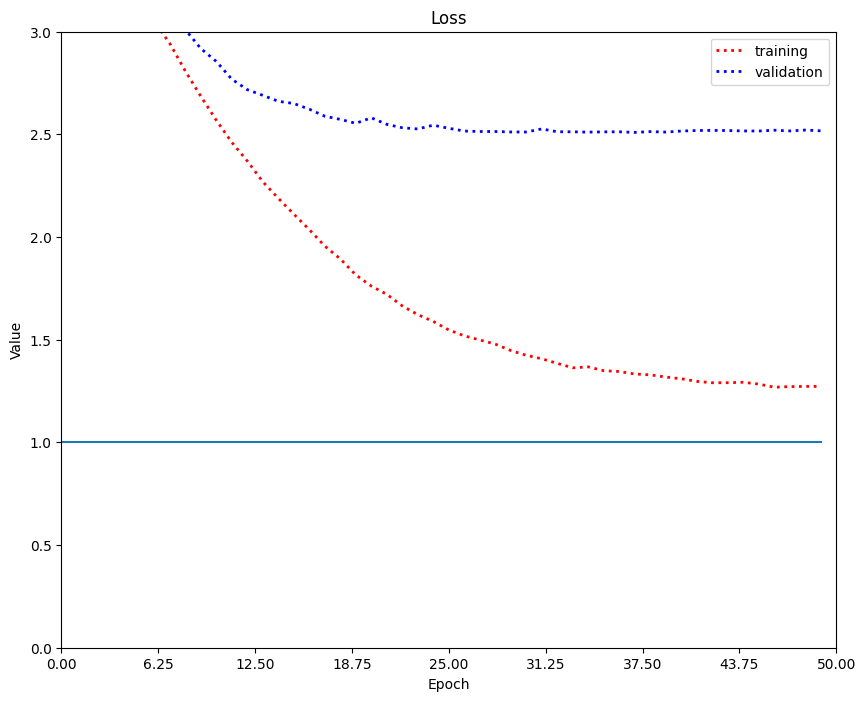}
    \caption{Graph of Scratch Model Loss Value from Indonesia Dataset Against Epoch}
    \label{indo_scratch_loss}
\end{figure}

Based on Figure \ref{indo_scratch_loss}, it's evident that the loss value of this Scratch model is quite high, exceeding 1. Compared to the loss values of the pre-trained models being tested, the range of loss values in this Scratch model is much wider. To minimize the loss, a more complex model, such as a pre-trained model, may be necessary.

\begin{figure}[H]
    \centering
    \includegraphics[width=7cm]{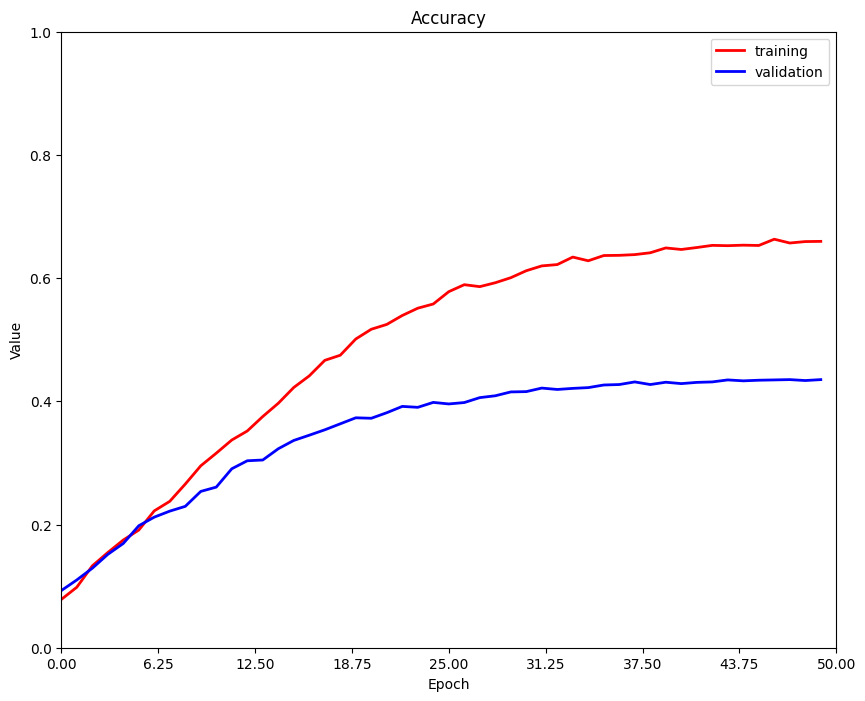}
    \caption{Graph of Scratch Model Accuracy Value from Indonesia Dataset Against Epoch}
    \label{indo_scratch_acc}
\end{figure}

Based on Figure \ref{indo_scratch_acc}, it's apparent that the accuracy value of this Scratch model is lower compared to the pre-trained models tested. Additionally, the range of accuracy values is considerably wider compared to the accuracy values of the other tested pre-trained models. To enhance the accuracy value of the Scratch model, a more complex model may be required.

\subsection{Discussion}
{\color{black} From the results of tests on the Vietnamese Herbal Plants Dataset and the Indonesian Herbal Plants Dataset, it becomes evident that if the Vietnamese dataset contains more images and classes, achieving a higher accuracy value becomes more challenging. This complexity arises due to the increased likelihood of images and classes exhibiting greater similarity. Among all the pre-trained models tested, it is apparent that the pre-trained ConvNeXt\cite{convnext} model, with an accuracy level ranging from 91\% to 92\%, possesses a model architecture more suitable for application in herbal plant image datasets compared to other pre-trained models. In contrast, the pre-trained Swin Transformer model is considered to exhibit the lowest level of accuracy compared to the other pre-trained models. It is noteworthy that Vision Transformer-based models require larger datasets than their convolutional counterparts to perform well \cite{lu2022bridging}\cite{paul2022vision}.

Additionally, compared to all the pre-trained models tested, the scratch model attains the lowest accuracy value. This is attributed to the scratch model commencing with random parameters, unable to leverage the richness of features derived from large datasets such as ImageNet.

The number of epochs also significantly influences the accuracy value. Employing an appropriate number of epochs proves beneficial in enhancing accuracy. Furthermore, the CNN model exhibits robustness in classifying limited herbal plant datasets, even when the imagery of the herbal plant dataset itself is intricate. However, achieving a commendable accuracy value necessitates the utilization of a fairly complex CNN model, as demonstrated by the effectiveness of transfer learning.}

\section{Conclusion}
The tests conducted using pre-trained CNN models underscored the necessity of employing complex models to achieve accurate herbal plant classification. Transfer learning proves to be a valuable approach in simplifying and expediting the creation of such complex models, thereby facilitating the production of accurate models for herbal plant classification.When transfer learning was applied to five specific pre-trained models - ResNet34, DenseNet121, VGG11, {\color{black}ConvNeXt, and Swin Transformer - the ConvNeXt pre-trained model yielded the best results. It achieved an accuracy rate of 92.8\% for the Vietnam Medicinal Plant Dataset and 92.5\% for the Indonesia Medicinal Plant Dataset. This highlights the effectiveness of leveraging transfer learning with ConvNeXt for herbal plant classification tasks.}

For future research endeavors, it is advisable to explore additional pre-trained models beyond the five examined in this study. Expanding the dataset of Indonesian herbal plants by increasing the number of classes would also enhance the diversity of plants available for classification. Given the rich variety of herbal plants in Indonesia, this would be particularly beneficial in capturing the full spectrum of herbal diversity. Moreover, it is worth considering the exploration of alternative Dynamic Learning Rate schedulers beyond the ExponentialLR scheduler used in this study. Experimenting with different schedulers could provide valuable insights into optimizing the training process and improving model performance.

\section*{Acknowledgement}
The authors would like to express gratitude to the Artificial Intelligence Center of Brawijaya University for providing a server.

\appendix


\bibliographystyle{ieeetr}
\bibliography{cas-refs}




\end{document}